\documentclass{article}

\usepackage{arxiv}

\usepackage[utf8]{inputenc} 
\usepackage[T1]{fontenc}    
\usepackage{hyperref}       
\usepackage{url}            
\usepackage{booktabs}       
\usepackage{amsfonts}       
\usepackage{nicefrac}       
\usepackage{microtype}      
\usepackage{graphicx}
\usepackage{natbib}
\usepackage{doi}

\usepackage{algorithm}
\usepackage{algorithmic}
\usepackage{subfiles}
\usepackage{tabularx} 
\usepackage{array}    
\usepackage{makecell} 
\usepackage{adjustbox}
\usepackage{multirow}
\usepackage{lscape}
\usepackage{subcaption}
\usepackage{amsmath}
\usepackage{cleveref}

\newcolumntype{Y}{>{\centering\arraybackslash}X}
\newlength{\extralength}
\newlength{\fulllength}
\title{Long-Term Forecasting of Multivariate Urban Data via Decomposition and Spatio-Temporal Graph Analysis}

\renewcommand{\shorttitle}{Long-Term Forecasting of Urban Data}


\author{ \href{https://orcid.org/0009-0003-5893-9715}{\includegraphics[scale=0.06]{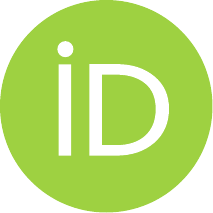}\hspace{1mm}Amirhossein Sohrabbeig} \\
	Department of Electrical and Computer Engineering\\
    University of Alberta\\
    Edmonton, AB T6G 1H9, Canada \\
	\texttt{sohrabbe@ualberta.ca} \\
    \And
    \href{https://orcid.org/0000-0002-6711-5502}{\includegraphics[scale=0.06]{orcid.pdf}\hspace{1mm}Omid Ardakanian} \\
	Department of Computing Science\\
    University of Alberta\\
    Edmonton, AB T6G 1H9, Canada \\
	\texttt{ardakanian@ualberta.ca} \\
    \And
    \href{https://orcid.org/0000-0002-7780-5048}{\includegraphics[scale=0.06]{orcid.pdf}\hspace{1mm}Petr Musilek} \\
	Department of Electrical and Computer Engineering\\
    University of Alberta\\
    Edmonton, AB T6G 1H9, Canada \\
	\texttt{pmusilek@ualberta.ca} \\
}

\date{}

\hypersetup{
pdftitle={Long-Term Forecasting of Multivariate Urban Data via Decomposition and Spatio-Temporal Graph Analysis},
pdfsubject={},
pdfauthor={Amirhossein Sohrabbeig, Omid Ardakanian, Petr Musilek},
pdfkeywords={Multivariate time-series forecasting, Spatio-temporal graph neural networks, Attention mechanism, Time-series decomposition},
}

\begin{document}
\maketitle

\begin{abstract}
Long-term forecasting of multivariate urban data poses a significant challenge due to the complex spatiotemporal dependencies inherent in such datasets. This paper presents DST, a novel multivariate time-series forecasting model that integrates graph attention and temporal convolution within a Graph Neural Network (GNN) to effectively capture spatial and temporal dependencies, respectively. To enhance model performance, we apply a decomposition-based preprocessing step that isolates trend, seasonal, and residual components of the time series, enabling the learning of distinct graph structures for different time-series components. Extensive experiments on real-world urban datasets---including electricity demand, weather metrics, carbon intensity, and air pollution---demonstrate the effectiveness of DST across a range of forecast horizons, from several days to one month. Specifically, our approach achieves an average improvement of 2.89\% to 9.10\% in long-term forecasting accuracy over state-of-the-art time-series forecasting models.
\end{abstract}


\section{Introduction}
Analyzing patterns and predicting trends in urban data, 
such as electricity demand~\cite{nti2020electricity}, 
traffic flow and air quality~\cite{huang2021pm2,du2024}, 
weather and climate variables~\cite{mouatadid2024subseasonalclimateusa}, and 
carbon intensity~\cite{maji2022carboncast}, 
is critical for a wide range of applications, including infrastructure planning,
disaster preparedness, and addressing social inequalities.
Urban data is typically multimodal and collected across large geographical areas.
Due to variations in sensor types and deployment environments,
and complex spatiotemporal dependencies within the data,
building accurate and transferrable forecasting models is a non-trivial task.

While multivariate time-series forecasting has been widely studied, 
achieving accurate long-term predictions, specifically over horizons of two to six weeks, 
remains a significant challenge~\cite{haoyietal-informer-2021}.
This difficulty arises from the accumulation of errors and shifting dynamics.
Furthermore, when data originates from multiple heterogeneous sensors, 
there is currently no universally effective model that consistently performs well across domains.
Physics-based dynamical models, where available, are inaccurate for long-term forecasting~\cite{mouatadid2024subseasonalclimateusa}.
Traditional statistical methods, recurrent neural networks, and transformer-based models 
often fail to capture the complex dependencies inherent in multivariate time-series data, 
leading to performance degredation on long-term forecasts. 

In recent years, foundational models have shown promise in time-series forecasting 
due to their ability to model complex temporal dependencies and handle multivariate data.
However, their success hinges on access to large-scale pretraining data~\cite{woo2024unified}, 
and they often require extensive fine-tuning to adapt to new domains, 
limiting their practical applicability. 

Graph Neural Networks (GNNs) offer a promising alternative by naturally capturing both temporal and spatial dependencies, thereby improving forecast accuracy in multivariate settings~\cite{wu2020connecting}.
However, the underlying graph structure is typically unknown and must be inferred from the data.
In this paper, we explore the use of GNNs to address the challenges of long-term forecasting of multivariate urban time-series data, 
with an emphasis on improving forecast accuracy. \textit{A key insight of our approach is that the latent graph structures corresponding to different time-series components (e.g., trend, seasonality) are often distinct}. 
This observation motivates the application of time-series decomposition prior to graph structure learning.

We propose a novel model that combines the Graph Attention Network (GAT)~\cite{velivckovic2017graph} for learning spatial dependencies among variates with a Temporal Convolutional Network (TCN)~\cite{bai2018empirical} to capture temporal dependencies within each variate. 
We construct a graph where nodes represent individual variates and edges encode dependencies inferred from historical data. 
Leveraging GATv2~\cite{brody2021attentive}, the model dynamically adjusts the influence of each node on neighboring nodes. To further enhance accuracy, we introduce a preprocessing step that decomposes the time-series into its fundamental components: trend, seasonal, and residual components. 
This enables the model to focus on the intrinsic structure of the data, 
using a distinct graph structure for each component.

We evaluate our model on multiple real-world urban datasets, 
demonstrating significant improvements in long-term forecasting performance over the state-of-the-art.
These results underscore the model's potential for practical deployment in diverse application domains.

The contributions of this paper are summarized below:
\begin{itemize}
    \item We propose a novel multivariate time-series forecasting model, called DST, that integrates GATv2 and TCN to learn the salient spatial and temporal features, thereby enhancing the accuracy of long-term forecasting.
    \item We implement a decomposition-based preprocessing step to isolate trend, seasonal, and residual components of the time-series, and learn a graph structure for each component independently. Together, they allow the model to tailor its predictions to distinct temporal dynamics, significantly improving its accuracy.
    \item We validate our model's effectiveness through extensive evaluation on four real-world datasets, showcasing on average 2.89\% to 9.10\% improvement over the state-of-the-art in time-series forecasting across these datasets.
\end{itemize}
We note that accurate forecasts across different types of urban data are often integral to decision-making tasks aimed at minimizing cost, energy consumption, and carbon emissions. For example, effectively distributing workloads across geographically dispersed datacenters requires simultaneous predictions of electricity demand, weather, and carbon intensity~\cite{DBLP:journals/corr/abs-2106-11750}.
Thus, even a slight improvement in the prediction accuracy in each dataset could translate to considerable savings in applications that rely on multiple such datasets.
This highlights the broader impact of our contributions.

\section{Related Work}
\label{sec:related_work}
Statistical methods have been the cornerstone of time-series forecasting for decades. 
Time-series models such as the Autoregressive Integrated Moving Average (ARIMA)~\cite{box1970distribution} and its variants have historically been used due to their simplicity and interpretability. 

Another widely used method is Exponential Smoothing (ETS)~\cite{holt2004forecasting, winters1960forecasting}, which assigns exponentially decreasing weights to past observations.
Despite their widespread use, these classical methods have notable limitations. They typically assume stationarity and linearity in the data, which can make them unsuitable for more complex, nonlinear, and large-scale multivariate time-series datasets. Furthermore, as the number of variables increases, these models often struggle with scalability and may overfit data.

Traditional machine learning models such as Support Vector Machines (SVM), Decision Trees, and Random Forests have been adapted to predict future values based on historical data. These models are capable of capturing nonlinear relationships better than statistical methods. Ensemble methods such as Gradient Boosting Machines (GBM)~\cite{friedman2001greedy} and XGBoost~\cite{chen2016xgboost} have been particularly successful in combining the predictions of multiple models to improve accuracy. However, these models often require significant feature engineering and may not fully capture the temporal dependencies inherent in time-series data.

Deep learning has revolutionized time-series forecasting 
by making it possible to learn nonlinear and complex temporal dependencies directly from the data. 
Early deep learning models for time-series forecasting 

focused on capturing long-term dependencies through recurrent neural network (RNN) architectures.
More recent advances include models such as N-BEATS~\cite{oreshkin2019n}, which employs backward and forward residual links and a deep stack of fully connected layers to achieve state-of-the-art performance on various benchmark datasets. Transformer-based models, such as Informer~\cite{haoyietal-informer-2021}, Autoformer~\cite{wu2021autoformer}, and Fedformer~\cite{zhou2022fedformer}, have also shown significant promise by leveraging self-attention mechanisms to capture long-range dependencies more effectively than RNN-based approaches. However, some recent studies show that transformer-based models may not always be the best fit for time-series forecasting~\cite{zeng2023transformers}.
Particularly, they may struggle with effectively capturing the inherent temporal patterns in time-series data due to their reliance on the self-attention mechanism, and have high computational overhead. DLinear and NLinear models~\cite{zeng2023transformers} take a simpler and more efficient approach that mitigates these issues by focusing on linear models that are more adept at handling the sequential nature of time-series data.

To address the high computational overhead of transformer-based models, PatchTST~\cite{nie2022time} segments time-series into patches, significantly improving computational efficiency and forecasting accuracy. Similarly, S-Mamba~\cite{wang2025mamba} employs a selective state-space approach to efficiently capture inter-variate correlations and temporal dependencies with nearlinear complexity, achieving state-of-the-art performance across several benchmarks.

Another recent work, CycleNet~\cite{lin2024cyclenet}, introduces Residual Cycle Forecasting~(RCF) to model periodic patterns using learnable recurrent cycles and to predict residual components. 
However, CycleNet assumes stable periodicities, making it less effective for dynamic or irregular cycle datasets, i.e., datasets with regularly repeating patterns such as daily or seasonal patterns. Additionally, it models each time-series channel independently, failing to capture interdependencies between variables, which can limit its performance on multivariate time-series data.

Recent advances in time-series forecasting feature foundation models such as TimesFM~\cite{das2023decoder} and MOIRAI~\cite{woo2024unified}, which generalize across diverse datasets without additional training. TimesFM, using a decoder-only architecture with input patching, achieves near state-of-the-art zero-shot performance. MOIRAI, with a masked encoder architecture, addresses challenges like cross-frequency learning and any-variate forecasting, leveraging the Large-scale Open Time-Series Archive~(LOTSA)~\cite{woo2024unified} for training. Both models demonstrate the potential of foundation models in handling complex and varied time-series forecasting tasks. However, these models can be computationally intensive, require significant resources for training and deployment, and often lack interpretability.

Graph Neural Networks~(GNNs) have emerged as a powerful paradigm for multivariate time-series forecasting by modeling variables as nodes and their interdependencies as edges in a graph. The MTGNN~\cite{wu2020connecting} framework introduces a joint architecture combining temporal and graph convolution modules, with a graph learning layer that adaptively constructs an adjacency matrix to uncover hidden variable relationships. In contrast, MTGODE~\cite{jin2022multivariate} employs neural ordinary differential equations over dynamically learned graphs to model continuous spatial-temporal dynamics, offering improved accuracy and robustness in forecasting tasks.

Despite the large amount of work that utilizes GNNs for time-series forecasting,
discovering optimal graph structures and integrating graph learning with time-series forecasting remains a significant challenge. 
While recent advances have introduced adaptive graph learning~\cite{9206739} and automated spatio-temporal fusion techniques~\cite{li2021spatialtemporalfusiongraphneural}, these methods come with their own limitations. Specifically, adaptive graph learning can be computationally expensive, and spatio-temporal fusion mechanisms may struggle to effectively capture long-range dependencies, leading to potential inaccuracies in forecasting~\cite{10.1145/3696661}.

To evaluate the proposed GNN-based long-term forecasting model, 
we consider the best model from each of the above categories---namely linear models, state-space models, transformer-based models, and foundation models---as our baseline.

\section{Methodology}
Let $X = (x^1, x^2, \ldots, x^T) \in \mathbb{R}^{T \times D}$ represent a time-series of $T$ observations, each being $D$-dimensional, i.e. emitted by $D$ sensors. Time-series forecasting concerns predicting future values based on past observations. Given a look-back window of length $l$ ending at time $t$, denoted as $X^{t,l} = (x^{t-l+1}, \ldots, x^{t-1}, x^t)$, the goal is to forecast the next $h$ steps of the time-series. This is achieved by learning the function $f_\theta(X^{t,l}) = Y^{t,h}=(y^{t+1}, \ldots, y^{t+h})$, where $\theta$ represents the parameters of the forecasting model. We refer to each pair of look-back and forecast windows as a sample. Since we focus on long-term forecasting, the forecast window is assumed to be long, usually several days to a few months.

We model the input and output time-series as a graph $G = (V, E)$. Each sensor $i$ is represented as a node $v_i$ that might have dependencies with other sensors. These dependencies are modeled using directed, weighted edges connecting the respective nodes. 
These connections are assumed to be static, i.e., the graph structure does not change over time. 

The node features are initially the historical sensor readings with the look-back window, and ultimately the forecast values.

Our methdology has three main steps:
\begin{enumerate}
    \item \textbf{Decomposition}: Decomposing the time-series into fundamental components (trend, seasonal, residual) to accurately capture inherent characteristics of each component.

    \item \textbf{Graph Structure Learning}: Learning the adjacency matrix $A\in \mathbb{R}^{D \times D}$ (with $D$ being the number of nodes) that captures the relationships between nodes for each time-series component. Each element of $A$, denoted as $a_{ij}$, represents the weight of the edge from node $v_i$ to node $v_j$.
    \item \textbf{Spatio-Temporal Forecasting via DST}: Utilizing the learned graph structure for each component, a GNN extracts spatio-temporal features for predicting future values of the respective component. The predicted future values of different components are then summed to produce the final forecast $Y^{t,h}$.
\end{enumerate}

Each of these steps will be explained in more detail in the subsequent sections.

\subsection{Decomposition}
\label{sec:decomposition}

The first step in our approach involves decomposing the time-series into its fundamental components: trend, residual, and one or more seasonal components. This decomposition is crucial as it isolates the inherent characteristics of the time-series, thus improving the accuracy of forecasting~\cite{forecast5040037}. 

The decomposition method adopted here is inspired by the MSTL decomposition technique~\cite{bandara2025mstl}.
It begins with the application of a weighted  moving average filter to the time-series, where the weights for this kernel are derived from the tricube function given below:
\begin{equation}
\text{tricube}(x) = 
\begin{cases}
(1 - |x|^3)^3, & \text{if } |x| \leq 1, \\
0, & \text{if } |x| > 1.
\end{cases}
\end{equation}
Intuitively, it assigns higher weights to nearby data points and lower weights to distant ones. The extracted trend is then subtracted from the original time-series to obtain the detrended series.

Next, the period of seasonality (e.g., daily or weekly patterns) must be determined. This can be done using various methods, such as autocorrelation analysis, Fourier transform, data visualization, etc. 

The next step is to capture seasonal patterns in the detrended series. A simple aggregation approach with a sliding window of appropriate length (denoted as $k$) and stride (denoted as $m$) is used for this purpose. 
The seasonal pattern is identified by averaging the segments covered by the sliding window. 
\begin{figure}[!t]
    \centering
    \includegraphics [width=0.5\textwidth]{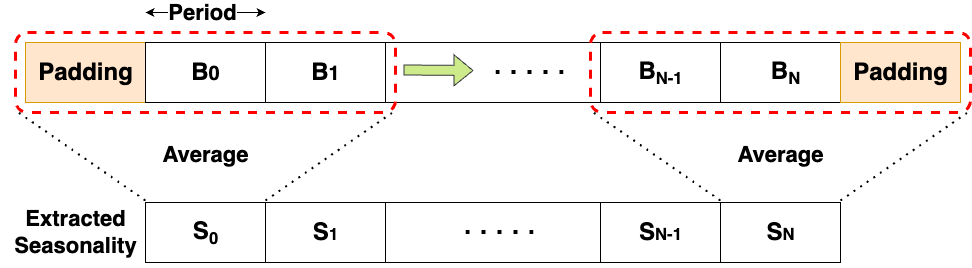}
    \caption{Extracting seasonal patterns using a sliding window of size $k$ and a stride of $m$ times the length of the seasonality period, respectively. 
    In this example, $k=3$ and $m=1$.}
    \label{fig:seasonality}
\end{figure}
Figure~\ref{fig:seasonality} illustrates this process. In each step of moving the sliding window, $S_i$ is calculated using the following equation:
\begin{equation}
\label{eq:seas}
    S_i = \frac{1}{k} \sum_{j=i-\lfloor k/2 \rfloor}^{i+ \lfloor k/2 \rfloor} B_i, 
\end{equation}
where $B_i$ is the i-th segment of the detrended time-series. Each segment contains $p$ data points, where $p$ represents the length of the seasonality period inferred in the previous step. Hence, the data points in $B_i$ have indices $i\times p$ to $(i+1)\times p-1$. To preserve the input length, padding is added to both ends of the time-series. This padding consists of copies of the first and last blocks.

Finally, the so-obtained seasonal components are subtracted from the detrended series to obtain the residual component.

\subsection{Graph Structure Learning}
\label{sec:graph_construction}

Since the underlying graph structure for each time-series component is not known in advance, we have to infer it from historical data. 
To this end, we first apply the decomposition process described in the previous section to the entire training dataset, then independently construct a graph for each resulting component. 
We note that graph structure learning is an offline task as the graph structures are assumed to be static.

The overall procedure is summarized in Algorithm~\ref{alg:graph_const}. It starts with decomposing the time-series into its fundamental components, where the number of these components is assumed to be identical across all variates.
To ensure computational efficiency, each component is first downsampled to a level that preserves its essential structural patterns.
To quantify similarity between sensor readings within each time-series component, we compute pairwise distances between the variates using Dynamic Time Warping (DTW)~\cite{muller2007dynamic}. DTW is selected for its robustness in capturing nonlinear temporal alignments between time-series. Each node is then connected to its $K$ nearest neighbors according to the DTW distances, where $K$ is a hyperparameter. 

\begin{algorithm}[tb]
\caption{Graph Structure Learning for Time-Series Components}
\label{alg:graph_const}
\begin{algorithmic}[1]
\REQUIRE time-series data $\{X_i\}_{i=1}^D$, number of nearest neighbors $K$
\ENSURE graph $G_i$ for each time-series component $i$
\FOR{each time-series variate $X_i$}
    \STATE $\hat{X}_i \gets \text{Detrend}(X_i)$
    \STATE $\{P_1, P_2, ..., P_n\} \gets \text{GetSeasnonalityPeriods}(\hat{X}_i)$
    \FOR{each seasonality period $P_j$ in $\{P_1, P_2, ..., P_n\}$}
        \STATE Compute seasonal component $S_j$
    \ENDFOR
    \STATE Compute the residual component
\ENDFOR

\FOR{each extracted component of $X$ denoted as $T_i$}
    \STATE Downsample $T_i$
    \STATE Build matrix $A_i$ by computing DTW between variates
    \STATE Construct $G_i$ by selecting top $K$ values in $A_i$
\ENDFOR

\end{algorithmic}
\end{algorithm}

\begin{figure}[!ht]
    \centering
    \begin{subfigure}[b]{0.23\textwidth}
        \centering
        \includegraphics[width=\textwidth]{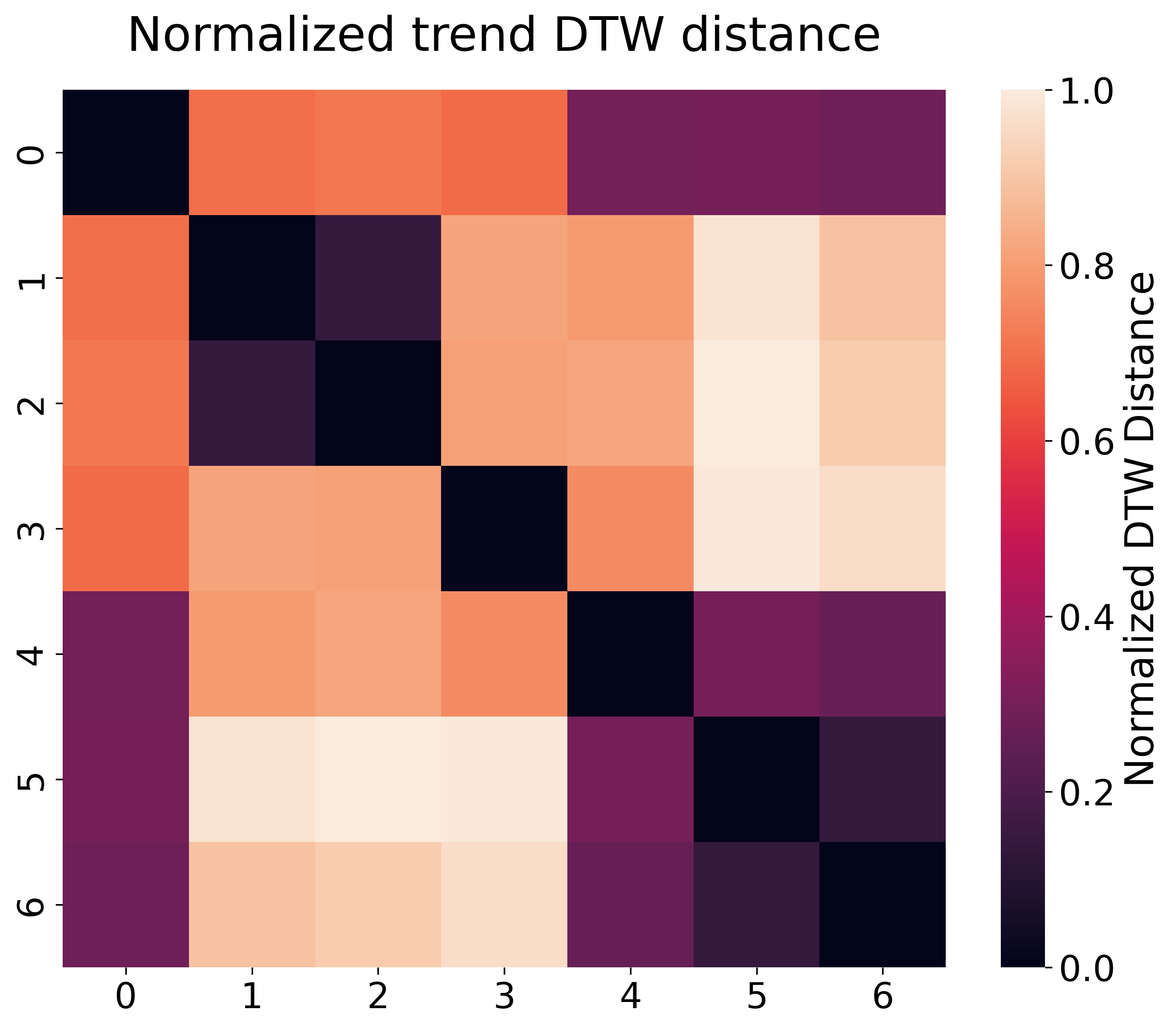}
        \caption{}
        \label{fig:heatmap1}
    \end{subfigure}
    \hspace{0.5em}
    \begin{subfigure}[b]{0.23\textwidth}
        \centering
        \includegraphics[width=\textwidth]{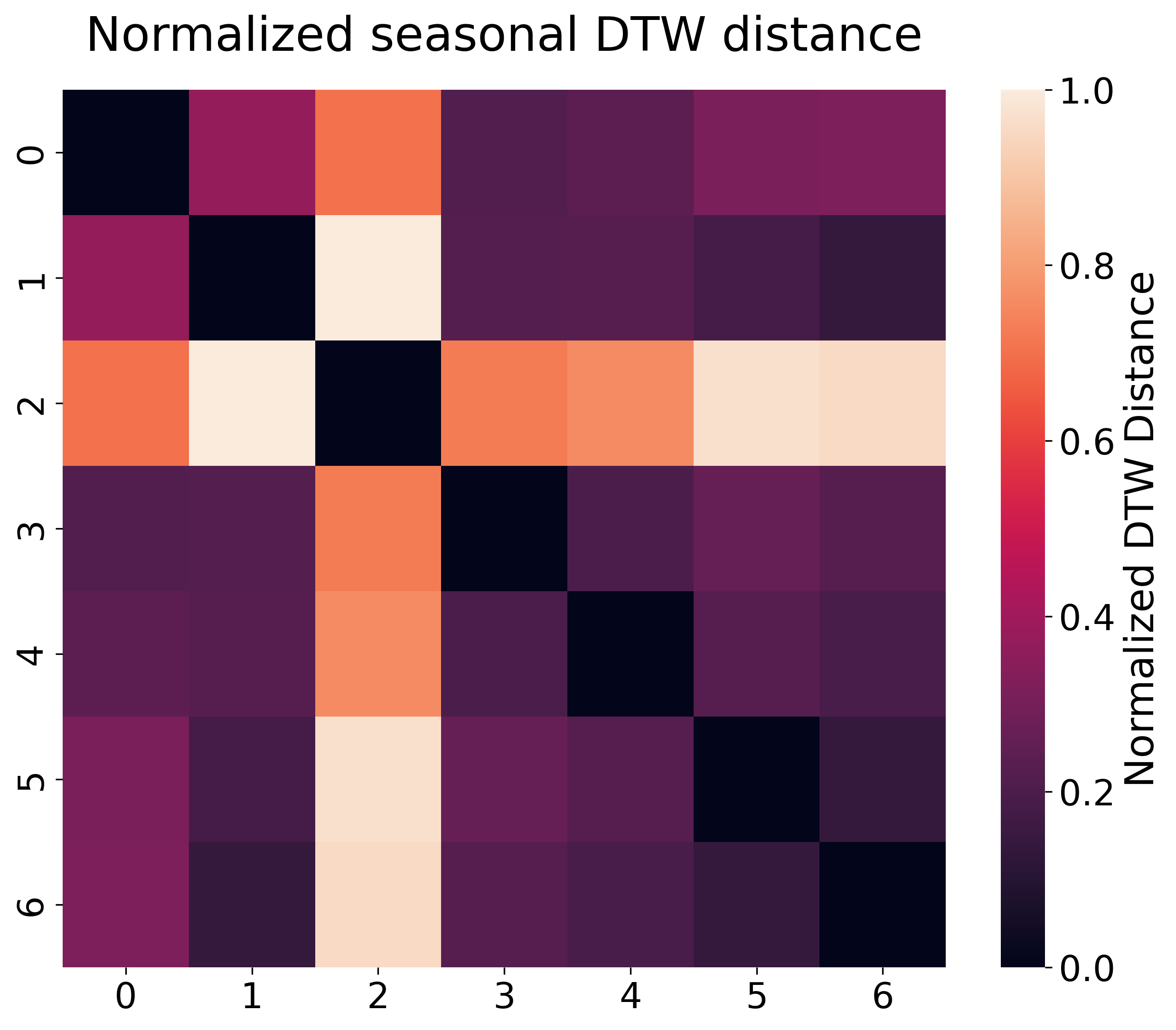}
        \caption{}
        \label{fig:heatmap2}
    \end{subfigure}
    \hspace{4em}
    \begin{subfigure}[b]{0.30\textwidth} 
        \centering
        \includegraphics[width=\textwidth]{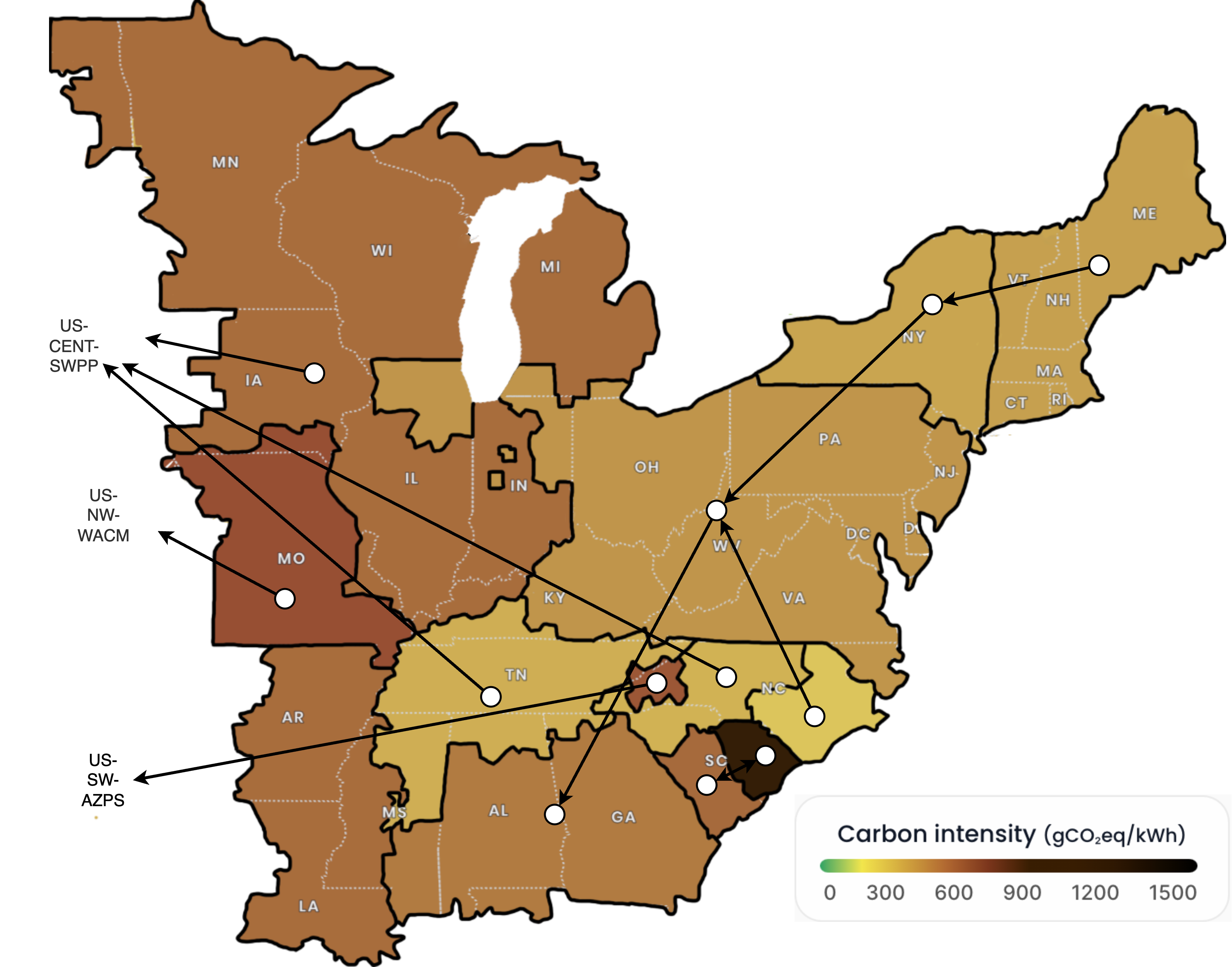}
        \caption{}
        \label{fig:USA}
    \end{subfigure}
    \caption{
        DTW distance heatmaps of (a) trend and (b) seasonal components of the air pollution dataset (Section \ref{sec:datasets}); lower values indicate greater similarity, with all values min-max normalized. (c) Carbon intensity for selected US states at a given time, where darker colors indicate higher intensity. Data come from multiple power generation companies with many-to-many links to states; some arrows extend beyond the displayed map to conneced areas not shown.
    }
    \label{fig:heatmap}
\end{figure}

Figures \ref{fig:heatmap1} and \ref{fig:heatmap2} illustrate DTW distances between different nodes for both the trend and seasonal components of the air pollution dataset, described in Section~\ref{sec:datasets}. These heatmaps reveal distinct patterns of similarity, with lower normalized DTW values indicating stronger dependencies between nodes. Thus, the graphs constructed for the trend and seasonal components have different structures.
Our ablation study, described in Section~\ref{sec:ablation}, confirms that this results in a significant improvement over a GNN-based forecasting model that does not perform decomposition or uses the same graph structure for all components.

Figure~\ref{fig:USA} depicts the graph constructed for the trend component of the carbon intensity dataset, described in Section~\ref{sec:datasets}. As it can be seen, the inferred graph contains  directed edges, as one variate might be dependent on the lagged version of the other one, but the opposite is not necessarily true.

\subsection{Spatio-Temporal Forecasting via DST}

\begin{figure*}[ht]
    \centering
    \includegraphics[width=\textwidth]{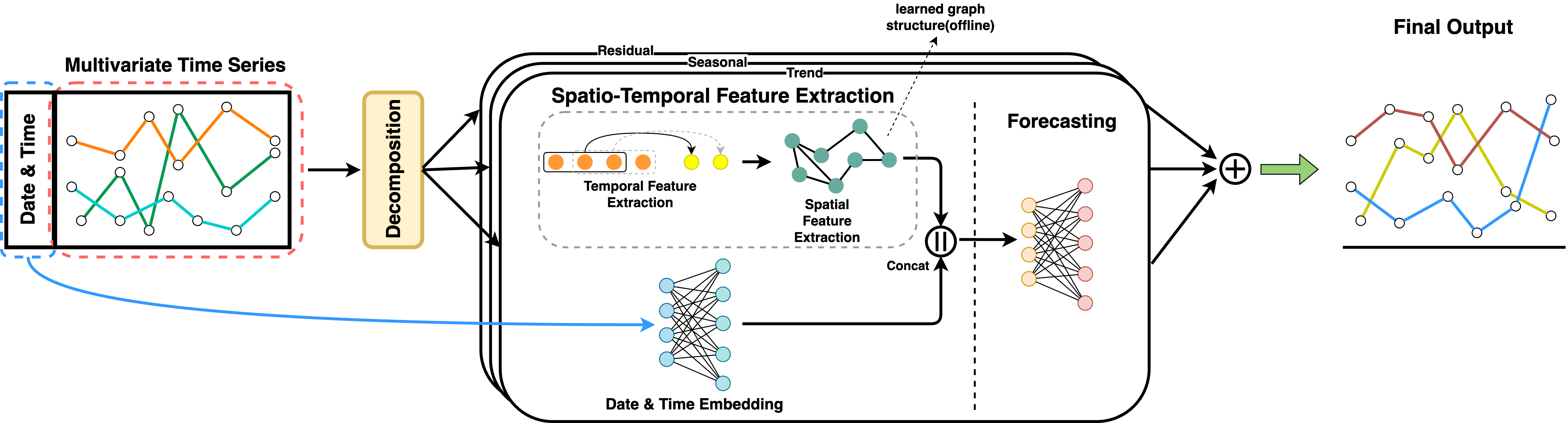}
    \caption{The overall architecture of proposed forecasting model. After decomposition, each component is handled independently.}
    \label{fig:Overal}
\end{figure*}

After constructing a graph for each time-series component in an offline fashion, we proceed with  spatio-temporal forecasting. The architecture of the proposed forecasting model, DST, is shown in Figure~\ref{fig:Overal}. The model's input is a sample of fixed-length historical values, $X^{t,l}$, and the corresponding date-time values, denoted as $X_\text{time}$.

\subsubsection{Sample Decomposition}
\begin{figure}[ht]
    \centering
    \includegraphics[width=0.5\linewidth]{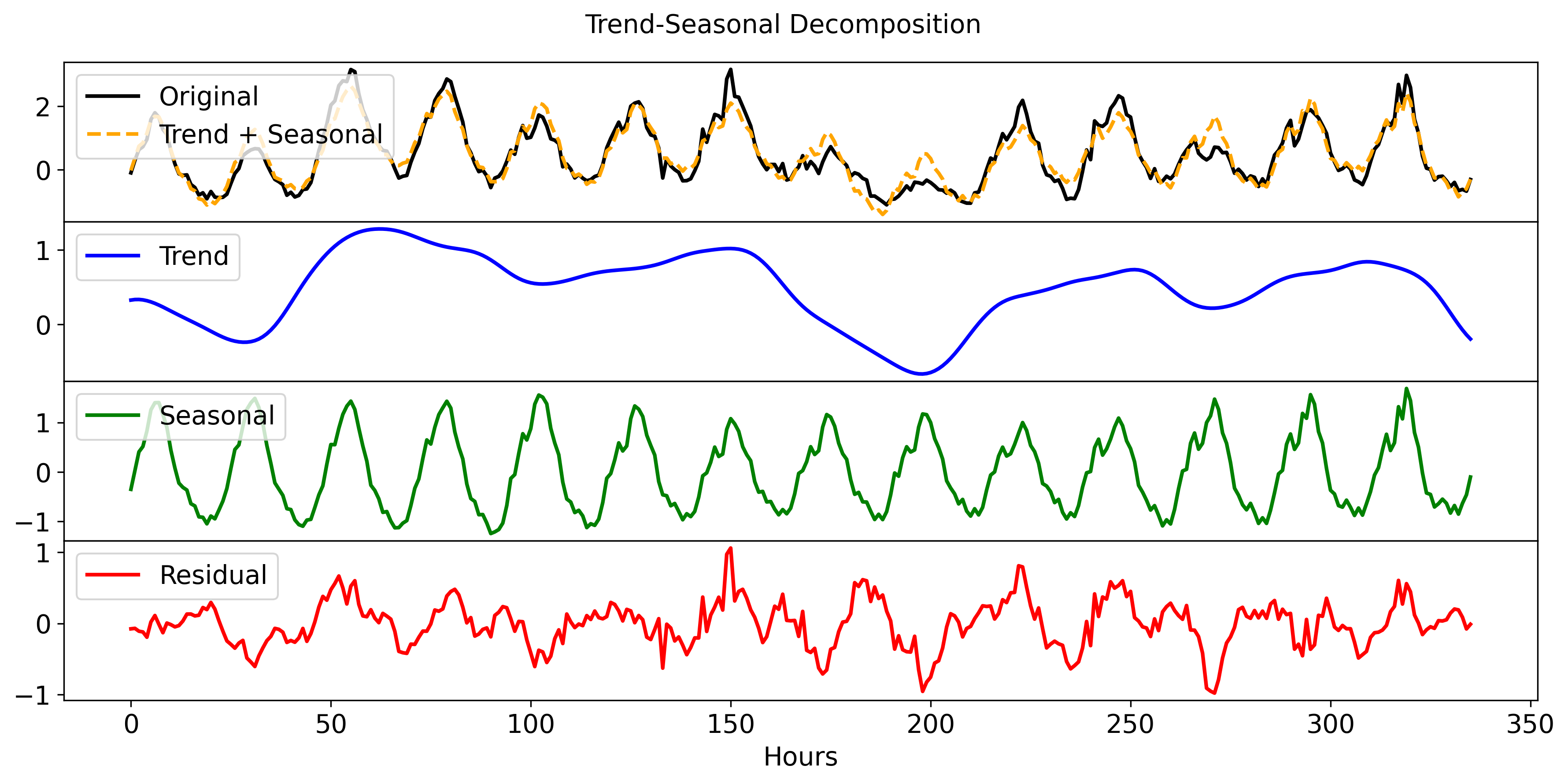}
    \caption{Decomposition of an electricity consumption sample over two weeks. The first subplot displays the original data alongside the sum of trend and seasonal components.}
    \label{fig:Trend_Seasonal_Decomposition}
\end{figure}

We begin by decomposing an input sample, without the date-time series, into its constituent components: trend, seasonality, and residual. For decomposition, we follow the same procedure described in Section~\ref{sec:decomposition}, but without  downsampling. 
A separate forecasting model is trained for each component using the graph structure learned for that component specifically. 
Figure~\ref{fig:Trend_Seasonal_Decomposition} illustrates the decomposition of a randomly selected sample from the Electricity dataset, described in Section~\ref{sec:datasets}.
In this case, there is just one seasonal component.

\subsubsection{Feature extraction}
\begin{figure}[!ht]
    \centering
    \includegraphics [width=0.5\textwidth]{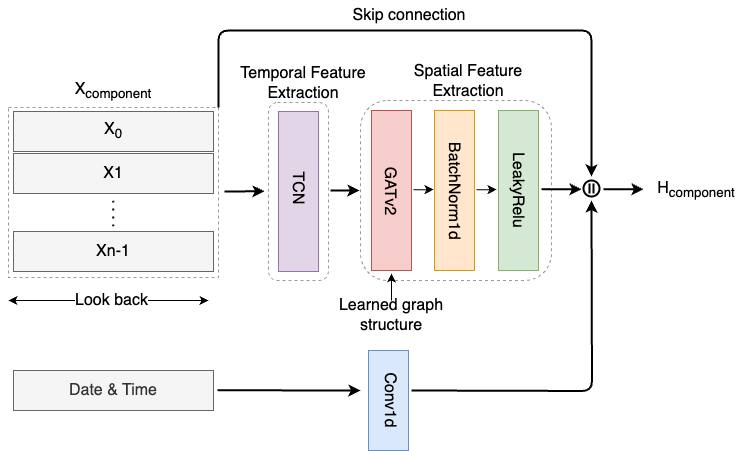}
    \caption{One layer of spatio-temporal feature extraction with time embedding. The input to this module is readings from $n$ sensors, $X_0$ to $X_{n-1}$. Each component from the decomposition module is processed separately using this layer. The spatial feature extractor is a GATv2 that relies on the learned graph structure.}
    \label{fig:spatiotemporal}
\end{figure}

The first module of DST is responsible for feature extraction. This module  handles each component, i.e., trend, seasonal, and residual, separately. 
Specifically, for each component, we extract and then combine three types of features, namely spatial and temporal features and date-time embedding.
The result is passed as input to the forecasting module, which generates the final output for that component.
Figure~\ref{fig:spatiotemporal} shows DST's feature extraction and forecasting modules.

To make the date-time embedding, we form a series ($X_\text{time}$) by putting the timestamps of input data points in a sequence. 
We then pass this to an encoder that uses 1D convolution to produce the date-time embedding.
Simultaneously, spatial and temporal features are extracted from the input data $X^{t,l}$. These temporal and spatial features are then combined and fed to the forecasting module as written below:
\begin{equation}\label{eq:combination}
    Y^{t,h} = f_{\text{forecast}}\left( f_{\text{time}}(X_\text{time}) \parallel f_{\text{spatio-temporal}}(X^{t,l}) \right),
\end{equation}
where $\parallel$ is the concatenation operator, $f_{\text{forecast}}$ is the forecasting model, $f_{\text{time}}$ is the date-time encoder, and $f_{\text{spatio-temporal}}$ is the spatiotemporal feature extractor, which is a combination of GATv2 and TCN as described next.

For spatial feature extraction, we utilize a GNN. In this model, each variate of the time-series is treated as a node, with its historical values within the look-back window serving as its features. The GNN requires a connection graph to understand the relationships between different nodes (variates) via message passing. The connection graph is inferred using the method described in Section~\ref{sec:graph_construction}. 

In a GNN, each node iteratively updates its state by aggregating information from its neighbors. The general formulation of a GNN layer is given below
\begin{equation}
\mathbf{h}_i' = \sigma \left( \mathbf{W} \mathbf{h}_i + \sum_{j \in \mathcal{N}_i} \mathbf{W}_j \mathbf{h}_j \right),
\label{eq:gnn}
\end{equation}
where $\mathbf{h}_i$ is the feature vector of node $i$, $\mathcal{N}_i$ represents the set of neighbors of node $i$ in the underlying graph, $\mathbf{W}$ and $\mathbf{W}_j$ are learnable weight matrices, and $\sigma$ is a nonlinear activation function. This aggregation allows the model to capture the dependencies between each node and its neighbors, which is crucial for spatial feature extraction.
The output $\mathbf{h}_i'$ is the updated  representation of node $i$, which is a weighted sum of its neighbors' representations. 
To increase learning capacity, we use the attention mechanism for aggregation. 
Specifically, we use
GATv2~\cite{brody2021attentive}, a dynamic graph attention variant of GAT,
to allow nodes to assign different amounts of importance to their neighbors in the aggregation step. This enables capturing time-varying dependecies.
The attention score $e_{ij}$ between node $i$ and its neighbor $j$ is computed as:
\begin{equation}
e_{ij} = \mathbf{a}^T \text{LeakyReLU} \left( \mathbf{W} [\mathbf{h}_i \parallel \mathbf{h}_j] \right)
\label{eq:gatv2}
\end{equation}

where $\mathbf{a}$ is a learnable weight vector, $\mathbf{W}$ is a weight matrix, $\mathbf{h}_i$ and $\mathbf{h}_j$ are the feature vectors of nodes $i$ and $j$, respectively, and $\parallel$ denotes concatenation. The attention scores are then normalized using Softmax:
\begin{equation}
\alpha_{ij} = \text{softmax}_j (e_{ij}) = \frac{\exp(e_{ij})}{\sum_{k \in \mathcal{N}_i} \exp(e_{ik})}
\label{eq:attention_scores}
\end{equation}
The final output representation of node $i$ is computed by passing a weighted sum of its neighbors' features to the activation function:
\begin{equation}
\mathbf{h}_i' = \sigma \left( \sum_{j \in \mathcal{N}_i} \alpha_{ij} \mathbf{W} \mathbf{h}_j \right)    
\label{eq:attention_final}
\end{equation}
This attention mechanism allows the GNN to focus on the most relevant neighbors, improving its ability to capture complex relationships in the graph.
We use just one GATv2 layer for spatial feature extraction.

For temporal feature extraction, we utilize Temporal Convolutional Networks (TCNs)~\cite{bai2018empirical}, which have demonstrated strong performance on sequential data across multiple domains. Our choice of TCN is motivated both by empirical results and by its architectural advantages over alternatives like LSTM and GRU.

TCNs capture temporal dependencies using convolutional operations over time steps, which offer parallelism during training, stable gradients, and a flexible receptive field. Compared to RNN-based models, such as LSTM and GRU, which process data sequentially and may struggle with long-range dependencies, TCNs enable faster training and are better suited to modeling long sequences, a key challenge in long-term time-series forecasting.

The adopted TCN has three core components~\cite{bai2018empirical}: causal convolutions, dilated convolutions, and residual connections.
Causal convolutions ensure that predictions at time $t$ depend only on past inputs, preserving the temporal order.

Dilated convolutions expand the receptive field without increasing model size, enabling the network to capture long-range patterns.

Residual connections promote efficient training of deeper models and help mitigate vanishing gradients.

Overall, TCNs offer a strong balance between expressiveness, training efficiency, and robustness for long-range temporal modeling, making them especially well-suited to the demands of multivariate, long-term forecasting of urban data.

\subsubsection{Forecasting}
Once features are extracted for each component of the time-series, we employ a linear forecasting model to produce forecasts for the entire forecast horizon at once; this approach is commonly referred to as \textit{direct multistep forecasting}. This choice is motivated by the demonstrated effectiveness of similar methods in previous studies~\cite{forecast5040037, zeng2023transformers}. Forecasts are generated simultaneously and independently for each component—trend, seasonal, and residual—mirroring the approach used in the feature extraction phase. Within each forecasting component, each variate is predicted in a univariate fashion. This means that only the spatial and temporal features extracted for each specific variate in the previous step are used for forecasting that variate. Nevertheless, these features are influenced by the features of other variates due to the way that they are extracted using a GNN.

The input to the forecasting model is the combination of spatio-temporal features and the time embedding (see Equation~\ref{eq:combination}). 

Figure~\ref{fig:spatiotemporal} shows how the input for the forecasting model is constructed.

We adopt a linear forecasting model that predicts values in the next $h$ time steps. 

The input to the model is a tensor $\mathbf{X}$ of shape $(B, D, l)$, where $B$ is the batch size, $D$ is the number of channels, and $l$ is the length of the input sequence.
During the forward pass, the model computes the prediction $\mathbf{Y}$ using the input tensor $\mathbf{X}$ and the learnable parameters. This is done through a linear transformation followed by the addition of a bias term:
\begin{equation}
\mathbf{Y} = \mathbf{W} \cdot \mathbf{X} + \mathbf{b},
\end{equation}
where $\mathbf{Y} = \mathbf{W} \cdot \mathbf{X}$ represents the Einstein summation operation defined as:
\begin{equation}
\mathbf{Y}_{bdh} = \sum_{l=1}^{L} \mathbf{W}_{dhl} \cdot \mathbf{X}_{bdl}
\end{equation}
for each batch $b$, channel $d$, and prediction time step $h$.
This process allows the model to learn a linear mapping from the input sequence to the predicted sequence, leveraging the entire input sequence to inform each prediction step. This implementation is more efficient than the method used by DLinear and NLinear~\cite{zeng2023transformers} models, which iteratively calculates forecasts for each variate.

\subsubsection{Training}
As explained earlier, DST takes advantage of a dedicated module for each component that was extracted via the decomposition method. While each module processes the respective component independently, all modules are jointly trained. For training, we used the PyTorch Lightning framework on a high-performance computing system featuring an AMD Ryzen 7 5800X 8-core processor (16 threads), 32GB RAM, and an NVIDIA GeForce RTX 3090 GPU with 24GB VRAM.

The model is optimized using the Adam optimizer with a learning rate of 0.0001 and a batch size of 32. Training is performed for a maximum of 20 epochs, with early stopping applied based on validation loss, using a patience of 3 epochs to mitigate overfitting. Gradient clipping with a threshold of 1.0 is employed to ensure training stability. The Mean Squared Error (MSE) loss function is used for training DST,  while both MSE and Mean Absolute Error (MAE) are reported during evaluation to provide a comprehensive assessment of forecasting performance. 

We note that the hyperparameter K, which defines the number of nearest neighbours for each node in the graph, is selected using grid search over a range spanning from 10\% to 100\% of all other nodes, based on validation performance.

\section{Evaluation}
We introduce the urban datasets and baseline models used for evaluating the long-term forecasting performance of DST.

\subsection{Datasets}\label{sec:datasets}

To comprehensively evaluate our model in realistic and diverse long-term forecasting scenarios, we utilize four multivariate time-series datasets spanning key domains: electricity usage, weather conditions, carbon intensity, and air pollution. These datasets were carefully selected to reflect a wide range of forecasting challenges, such as high dimensionality, varying temporal resolutions, and complex inter-variable dependencies.
As a result, they are suitable benchmarks for assessing the generalizability and robustness of multivariate time-series forecasting models.

We describe each dataset below:
\begin{itemize}
    \item \textbf{Electricity}\footnote{\url{https://archive.ics.uci.edu/ml/datasets/ElectricityLoadDiagrams20112014}}:
    This dataset consists of hourly electricity consumption data from 321 customers over a period of three years. It serves as a benchmark for modeling periodic, high-dimensional energy demand, where capturing temporal patterns and seasonal cycles is critical.

    \item \textbf{Weather}\footnote{\url{https://www.bgc-jena.mpg.de/wetter/}}:
    This dataset includes 21 meteorological metrics collected at 10-minute intervals throughout 2020 from a weather station at the Max Planck Biogeochemistry Institute. Its high-frequency nature makes it ideal for evaluating models' sensitivity to rapid fluctuations and short-term dependencies.

    \item \textbf{Carbon Intensity}\footnote{\url{https://www.electricitymaps.com/data-portal/united-states-of-america}}: This dataset contains hourly carbon intensity values recorded across 50 U.S. states and four territories over three years, this dataset introduces spatio-temporal diversity and is crucial for applications involving environmental sustainability and energy emissions forecasting.

    \item \textbf{Air Pollution}\footnote{\url{https://archive.ics.uci.edu/dataset/381/beijing+pm2+5+data}}:
    This dataset provides hourly PM2.5 concentrations recorded by the U.S. Embassy in Beijing, along with concurrent meteorological data from the Beijing Capital International Airport~\cite{Liang2015AssessingBP}. It presents a complex, multivariate forecasting task influenced by both environmental and anthropogenic factors.
\end{itemize}

These datasets collectively span different temporal granularities, input complexities, and domain characteristics, making them an effective testbed for evaluating models intended for urban forecasting. Moreover, they align with benchmarks used in recent foundational and graph-based forecasting research~\cite{zeng2023transformers, wang2025mamba}, ensuring relevance and comparability.
Table~\ref{tab1} summarizes the key characteristics of each dataset. 
There are no missing values in these datasets,
except for the air pollution dataset, which required imputation using averages of similar temporal contexts (same month, weekday, and hour). As a preprocessing step, we standardized
all features.

\begin{table}[!t]
\small
\setlength{\tabcolsep}{10pt} 
\centering
\begin{tabular}{l c c c}
\toprule
\textbf{Dataset} & 
\textbf{\# Variates} &
\textbf{Length} &
\textbf{Time Resolution} \\ 
\midrule
Electricity &321 &26,304 &1h \\ 
Air Pollution &7 &34,970 &1h \\
Carbon Intensity &49 &26,280 &1h \\ 
Weather &21 &52,696 &10min \\ 
\bottomrule
\end{tabular}
\caption{Statistical details of the real-world datasets used in this study.}
\label{tab1}
\end{table}

\subsection{Baseline Forecasting Models}
Following the literature review provided in Section~\ref{sec:related_work}, we select representative state-of-the-art baselines across multiple forecasting paradigms for comprehensive evaluation of our approach. Transformer-based models include Autoformer~\cite{wu2021autoformer} and PatchTST~\cite{nie2022time}, both of which are designed to effectively capture temporal patterns through attention mechanisms. Autoformer is chosen over early adaptations of transformer-based models for time-series forecasting, namely Informer~\cite{haoyietal-informer-2021} and Fedformer~\cite{zhou2022fedformer}, primarily due to its consistently stronger performance on our datasets, making it a more suitable representative of that class of models.

From the class of linear models, we incorporate DLinear~\cite{zeng2023transformers}, known for its efficiency and competitive performance on long-sequence forecasting tasks.

To assess the capabilities of large-scale pretrained models, we consider TimesFM~\cite{das2023decoder}, evaluated in a zero-shot inference setting without fine-tuning, thereby highlighting the model's generalization ability. MOIRAI~\cite{woo2024unified} is excluded from our evaluation as it is primarily designed for probabilistic forecasting, whereas our setup focuses on point forecasting.

Additionally, we evaluate against Simple-Mamba (S-Mamba)~\cite{wang2025mamba}, a recent selective state-space model that leverages a lightweight architecture to efficiently capture both inter-variate correlations and temporal dependencies with near-linear computational complexity.

We also include \textit{Repeat Last}, a naive baseline that simply repeats the last observed value. This serves as a reference for assessing the forecasting performance of other models.

Classical statistical models such as ARIMA and seasonal ARIMA (SARIMA) are excluded from direct comparisons due to their high computational cost for parameter selection and their consistently poor performance on long sequence time-series forecasting tasks, as reported in prior studies~\cite{wu2021autoformer, haoyietal-informer-2021}.
Moreover, we do not consider models that have been specifically designed for a particular domain or dataset (e.g. air pollution or carbon intensity) as part of our baseline comparisons. This is because they cannot be readily used in other domains, often requiring changes to their architecture or pre-processing steps.

\section{Results}
We now evaluate the performance of our model (labelled DST in Table~\ref{tab2}) and compare it with strong baselines in three different settings. In all experiments, the length of the look-back window is 336, and the learning rate is set to 0.0001. The length of the forecast horizon is specified in the second column of Table~\ref{tab2}. For our model, we decided to extract only one seasonal component due to the short length of the input. For instance, when the length of the look-back window is 336 hours (equivalent to two weeks), it is logical to extract daily seasonal patterns since multiple daily cycles are present in this period while extracting weekly patterns would be less meaningful considering the limited number of weekly cycles.

We consider 4 long-term forecast horizons, namely 96, 192, 336, and 720, as they are commonly used in long-term time-series forecasting studies~\cite{wu2021autoformer, zhou2022fedformer, zeng2023transformers}. These multi-day to multi-week horizons are also more suitable for downstream applications, such as generator scheduling, carbon-aware computing~\cite{maji2022carboncast}, and city-scale planning~\cite{mouatadid2024subseasonalclimateusa}, where long-range forecasts support critical decision-making tasks more effectively than short-term ones.

\begin{table*}[!ht]
\scriptsize 
\setlength{\tabcolsep}{4.8pt}
\begin{adjustbox}{width=\textwidth,center}
\begin{tabularx}{\textwidth}{p{6em} >{\centering\arraybackslash}p{7em} *{16}{c}}
\toprule
 \multicolumn{1}{c}{\multirow{2}{6em}{\textbf{Dataset}}} 
 & \multicolumn{1}{c}{\multirow{2}{6em}{\textbf{Look-ahead (\# samples)}}}
 & \multicolumn{2}{c}{\textbf{DST}}
 & \multicolumn{2}{c}{\textbf{s-Mamba}}
 & \multicolumn{2}{c}{\textbf{DLinear}}
 & \multicolumn{2}{c}{\textbf{Autoformer}}
 & \multicolumn{2}{c}{\textbf{TimesFM}} 
 & \multicolumn{2}{c}{\textbf{PatchTST}}
 & \multicolumn{2}{c}{\textbf{Repeat Last}}\\
\cmidrule(lr){3-4} \cmidrule(l){5-6} \cmidrule(l){7-8}  \cmidrule(l){9-10}  \cmidrule(l){11-12} \cmidrule(l){13-14} \cmidrule(l){15-16}
 & & \textbf{MSE} & \textbf{MAE} & \textbf{MSE} & \textbf{MAE} & \textbf{MSE} & \textbf{MAE} & \textbf{MSE} & \textbf{MAE} & \textbf{MSE} & \textbf{MAE} & \textbf{MSE} & \textbf{MAE} & \textbf{MSE} & \textbf{MAE} \\

\midrule
\multirow{4}{4em}{Air pollution}
&96     &\textbf{0.864}  &0.703     &0.874  &\textbf{0.699}     &\underline{0.872}  &\underline{0.701}     &0.947  &0.714     &0.914  &0.695     &0.901  &0.699     &1.579  &0.869
\\
&192    &\textbf{0.903}  &\textbf{0.719}     &\underline{0.926}  &\underline{0.728}     &0.931  &0.732     &1.008  &0.744     &0.973  &0.739     &0.944  &0.735     &1.725  &0.926     \\
&336    &\textbf{0.916}  &\textbf{0.722}     &\underline{0.954}  &\underline{0.738}     &0.955  &0.745     &1.057  &0.792     &0.969  &0.744     &0.958  &0.753     &1.799  &0.948     \\
&720    &\textbf{0.945}  &\textbf{0.735}     &\underline{0.986}  &\underline{0.746}     &1.015  &0.783     &1.063  &0.814     &1.009  &0.751     &0.989  &0.803     &1.877  &0.977     \\

\midrule
\multirow{4}{4em}{Carbon}
&96     &\textbf{0.232}  &\textbf{0.370}     &\underline{0.235}  &\underline{0.375}     &0.244  &0.381     &0.386  &0.506     &0.418  &0.561     &0.349  &0.497      &1.771  &1.039    \\
&192    &\textbf{0.248}  &\textbf{0.389}     &\underline{0.273}  &\underline{0.406}     &0.284  &0.412     &0.371  &0.491     &0.445  &0.594     &0.413  &0.569     &1.853  &1.077     \\
&336    &\textbf{0.266}  &\textbf{0.410}     &0.307  &0.430     &\underline{0.305}  &\underline{0.427}     &0.454  &0.532     &0.457  &0.612     &0.435  &0.588     &1.919  &1.106     \\
&720    &\textbf{0.282}  &\textbf{0.425}     &0.368  &0.476     &\underline{0.364}  &\underline{0.469}     &0.407  &0.502     &0.551  &0.687     &0.512  &0.617     &2.025  &1.141     \\

\midrule
\multirow{4}{4em}{Electricity}
&96     &\textbf{0.148}  &\textbf{0.273}     &\underline{0.166}  &0.292     &0.169  &0.293     &0.194  &0.309     &0.200  &\underline{0.291}     &0.198  &0.311     &1.221  &0.858     \\
&192    &\textbf{0.176}  &\textbf{0.298}     &\underline{0.189}  &\underline{0.312}     &0.191  &0.320     &0.211  &0.319     &0.209  &0.319     &0.214  &0.326     &1.223  &0.857     \\
&336    &\textbf{0.188}  &\textbf{0.310}     &0.209  &0.329     &\underline{0.204}  &\underline{0.321}     &0.218  &0.330     &0.211  &0.334     &0.237  &0.349     &1.297  &0.881     \\
&720    &\textbf{0.214}  &\textbf{0.335}     &0.252  &0.371     &0.255  &0.374     &\underline{0.240}  &\underline{0.351}     &0.258  &0.359     &0.284  &0.397     &1.369  &0.908     \\

\midrule
\multirow{4}{4em}{Weather}
&96     &\textbf{0.096}  &\textbf{0.206}     &0.127  &0.234     &0.131  &0.237     &\underline{0.109}  &\underline{0.212}     &0.112  &0.243     &0.115  &0.256     &0.130  &0.262     \\
&192     &\textbf{0.378}  &\textbf{0.416}     &0.412  &0.425     &0.423  &0.427     &\underline{0.401}  &\underline{0.419}     &0.439  &0.461    &0.437  &0.459     &0.451  &0.478     \\
&336     &\textbf{0.780}  &\textbf{0.604}     &0.834  &0.613     &0.845  &0.621     &\underline{0.806}  &\underline{0.605}     &0.812  &\underline{0.605}     &0.829  &0.618     &0.844  &0.620     \\
&720    &\textbf{1.237}  &\textbf{0.797}     &1.408  &0.825     &\underline{1.392}  &0.818     &1.398  &\underline{0.816}     &1.418  &0.876     &1.378  &0.133      &1.470  &0.934     \\

\bottomrule
\end{tabularx}
\end{adjustbox}
\caption{Comparison of the forecasting accuracy of DST and baselines on four dataset in the multivariate input, univariate output setting. The results were averaged over five runs to mitigate the effect of randomness. The best results are highlighted in \textbf{bold} and the second-best results are \underline{underlined}.} 

\label{tab2}
\end{table*}

As shown in Table~\ref{tab2}, DST achieves the best performance, with respect to MSE and MAE, in almost all settings.
Notably, it outperforms the best baseline model, which is s-Mamba in most cases, by up to 9.10\% on average.
It can be seen that the gap between DST and the second best method generally widens for longer forecast horizons.
This highlights the strength of the proposed framework for long-term multivariate time-series forecasting.

\begin{figure*}[!ht]
    \centering
    \begin{subfigure}[b]{0.24\textwidth}
        \centering
        \includegraphics[width=\textwidth]{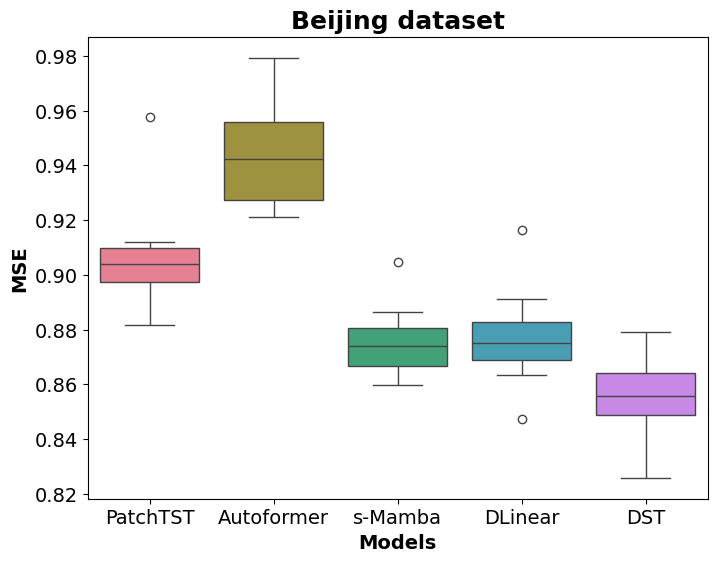}
        \label{fig:box_plot1}
    \end{subfigure}
    \hfill
    \begin{subfigure}[b]{0.24\textwidth}
        \centering
        \includegraphics[width=\textwidth]{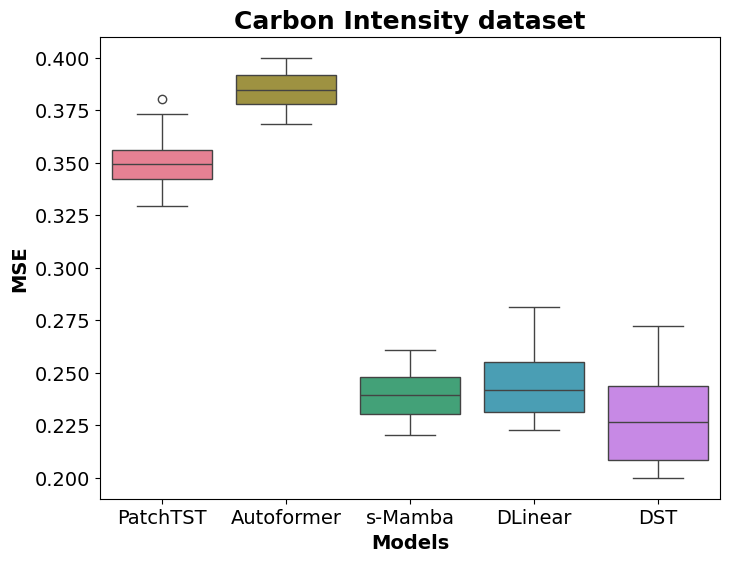}
        \label{fig:box_plot2}
    \end{subfigure}
    \hfill
    \begin{subfigure}[b]{0.24\textwidth}
        \centering
        \includegraphics[width=\textwidth]{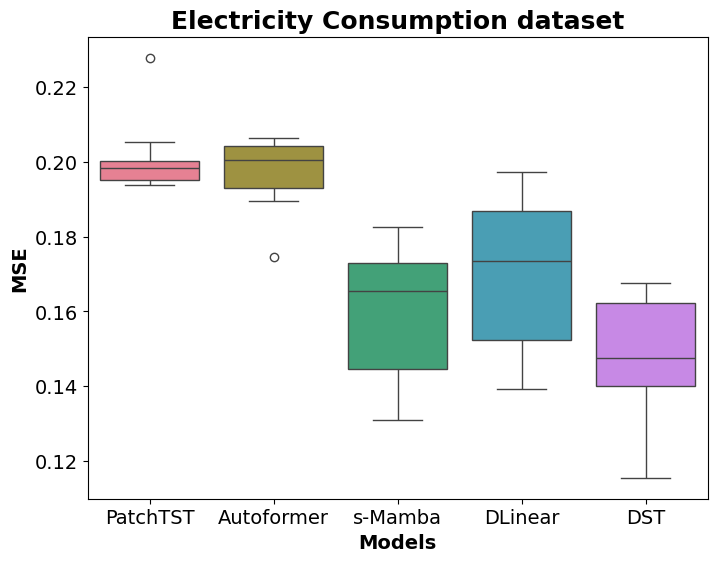}
        \label{fig:box_plot3}
    \end{subfigure}
    \hfill
    \begin{subfigure}[b]{0.24\textwidth}
        \centering
        \includegraphics[width=\textwidth]{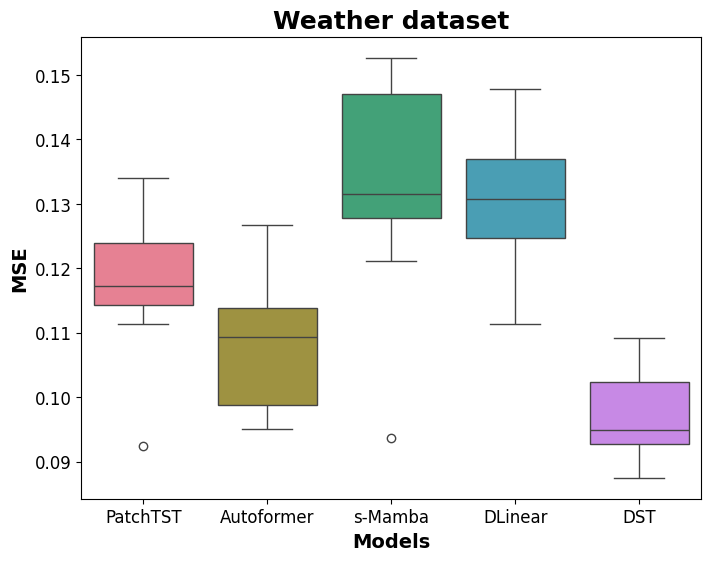}
        \label{fig:box_plot4}
    \end{subfigure}
    \caption{MSE comparison of our proposed model against baselines on different datasets. This is a multivariate input, multivariate output task, with input and output lengths of 336 and 96, respectively. Note that the axis scales are exaggerated for better visualization of the differences between models.}
    \label{fig:four_box_plots}
\end{figure*}

\begin{figure*}[!ht]
    \centering
    \begin{subfigure}[b]{0.48\textwidth}
        \centering
        \includegraphics[width=\textwidth]{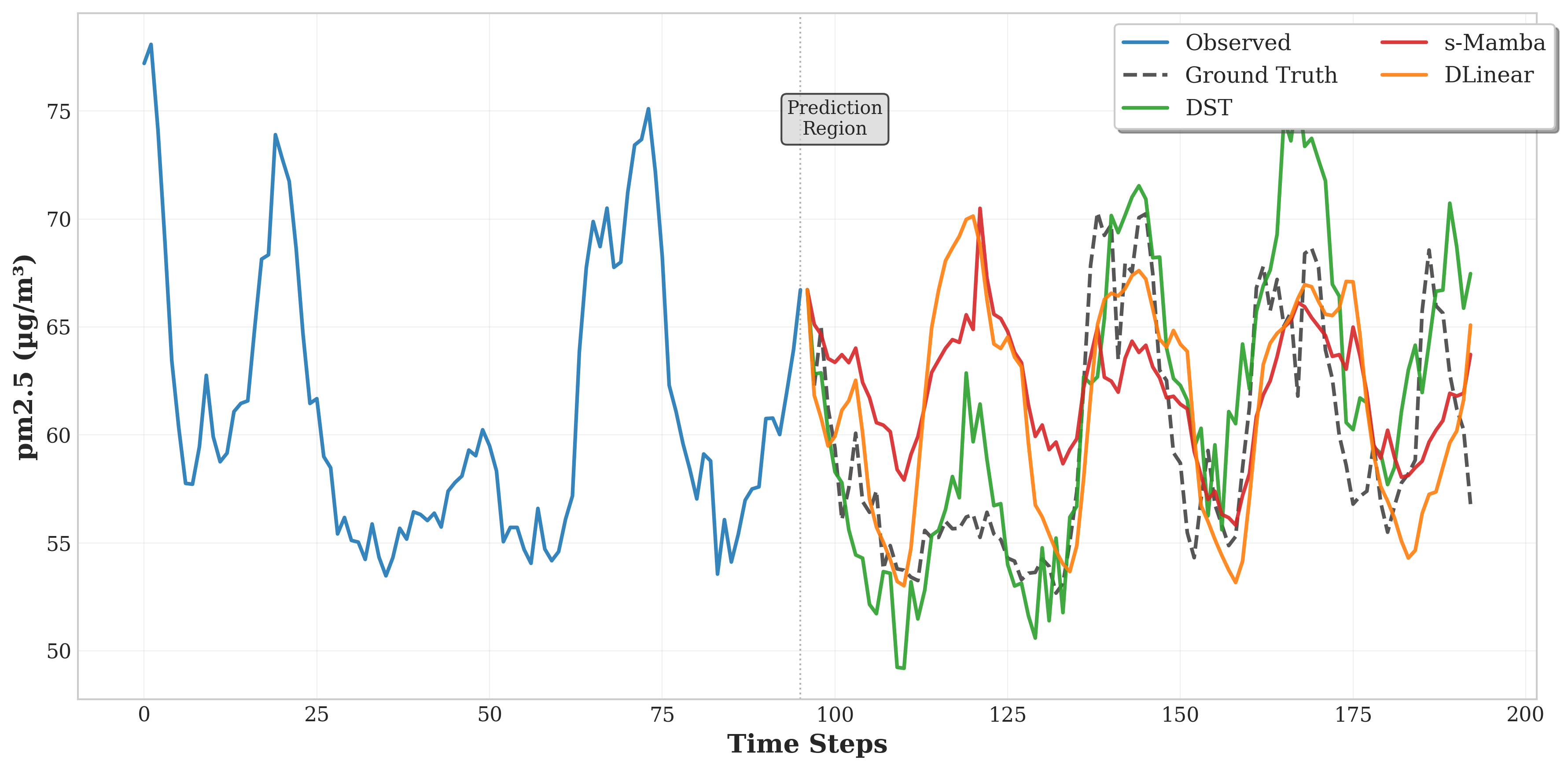}
        \caption{Air Pollution}
    \end{subfigure}
    \hfill
    \begin{subfigure}[b]{0.48\textwidth}
        \centering
        \includegraphics[width=\textwidth]{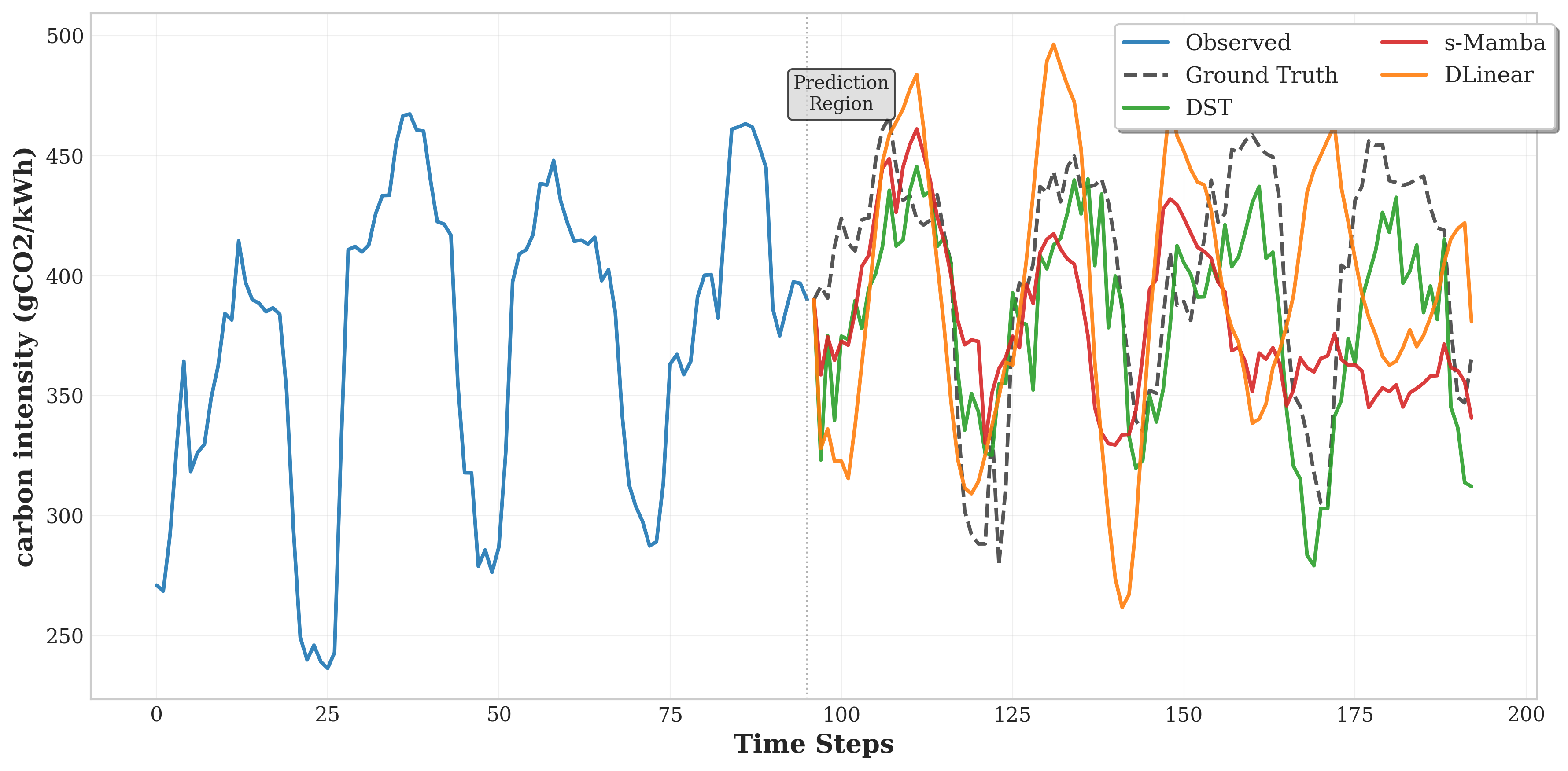}
        \caption{Carbon Intensity}
    \end{subfigure}
    \vskip\baselineskip
    \begin{subfigure}[b]{0.48\textwidth}
        \centering
        \includegraphics[width=\textwidth]{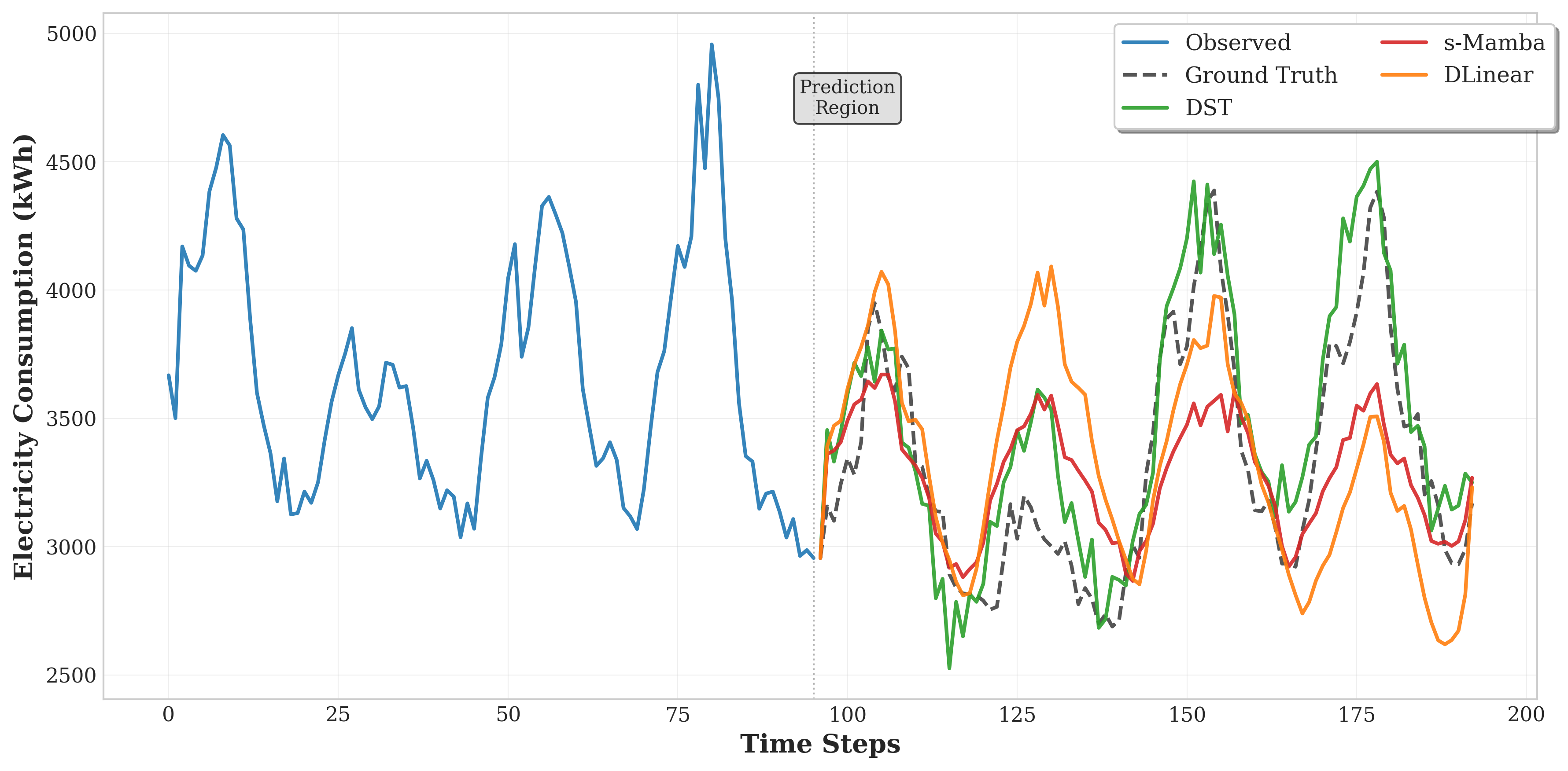}
        \caption{Electricity}
    \end{subfigure}
    \hfill
    \begin{subfigure}[b]{0.48\textwidth}
        \centering
        \includegraphics[width=\textwidth]{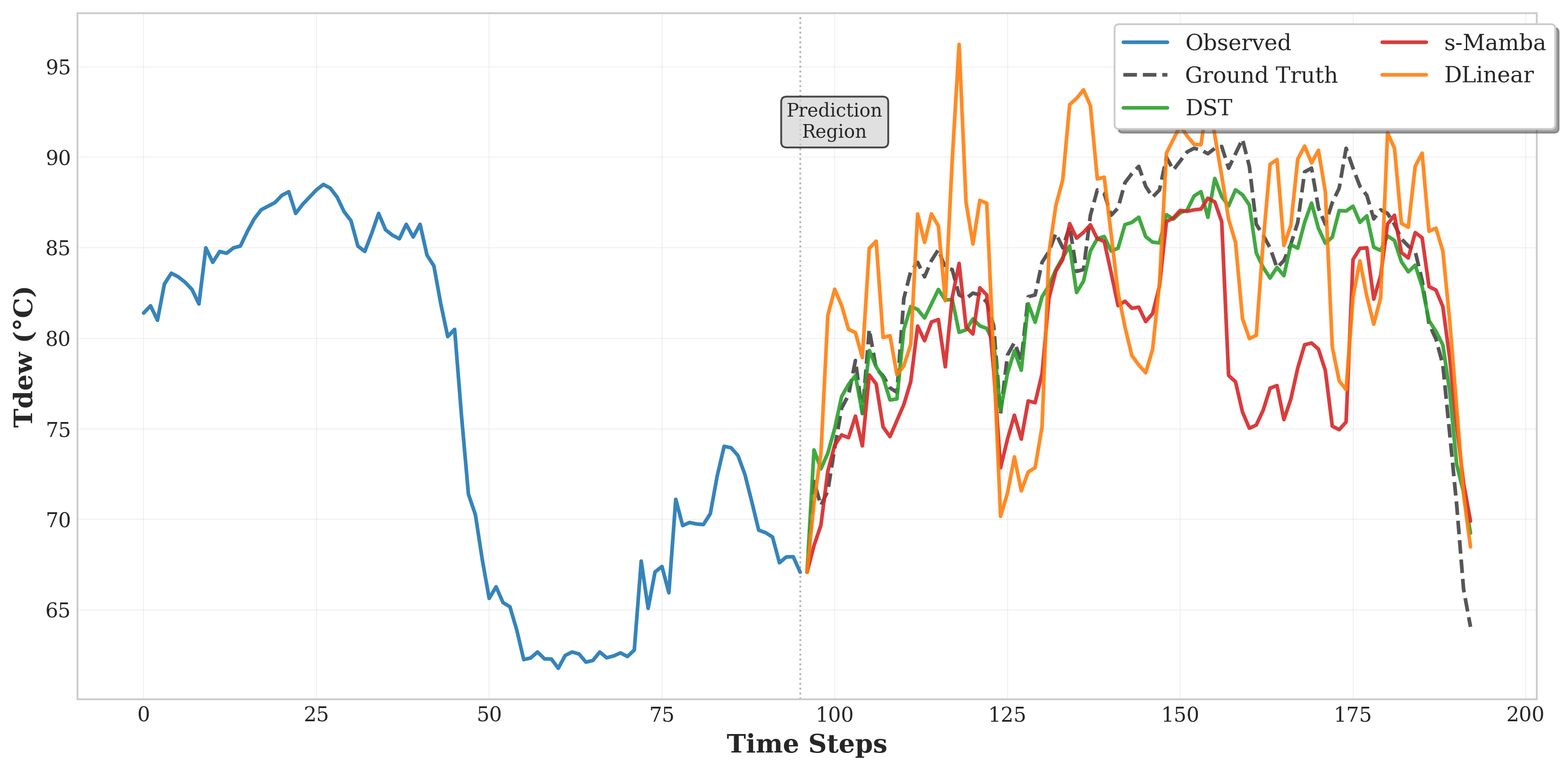}
        \caption{Weather}
    \end{subfigure}
    \caption{Comparison between the forecasts of DST, and two best performing baselines: s-Mamba and DLinear, on selected samples from a randomly chosen variate in each dataset.}\label{fig:prediction_comparison}
\end{figure*}

To further demonstrate the effectiveness of DST, we provide a visual comparison of its predictions against those of the best-performing baselines, s-Mamba and DLinear, in Figure~\ref{fig:prediction_comparison}.
These examples highlight the ability of our model to more accurately capture the temporal dynamics of the target time-series across different domains. In particular, DST demonstrates smoother forecasts and better alignment with the ground truth values. These qualitative results complement the quantitative improvements reported in Table~\ref{tab2}, offering additional insight into the predictive strengths of our approach.

Apart from s-Mamba, DLinear also exhibits reasonable performance among the baselines. 
Despite its good performance, it is a univariate model that lacks a mechanism to extract spatial dependencies. This limitation becomes evident in the multivariate input and univariate forecasting setting, where forecasting a variate may benefit from capturing lagged dependencies with other variates. Univariate models cannot leverage this information, leading to inferior performance compared to the models that can, such as our proposed model (DST) and transformer-based models. This is particularly evident in the forecasting of PM2.5 air pollution dataset.

Transformer-based forecasting models, such as Autoformer, Fedformer, and Informer, are highly complex and over-parameterized, making them susceptible to overfitting. This issue is apparent in the forecasting of carbon intensity. Additionally, we have included the forecasting performance of TimesFM, which was obtained without fine-tuning (i.e. its zero-shot performance, which was also reported in~\cite{das2023decoder}). This lack of fine-tuning likely contributes to its subpar performance, as it might not have adapted well to the specific characteristics of each dataset. Pretrained models must typically undergo additional training to align with the unique temporal patterns and distributions of new datasets. Without this adaptation, performance may suffer in comparison to models trained directly on the target datasets.

Figure~\ref{fig:four_box_plots} presents a comparative analysis of the forecasting performance, in terms of the distribution of MSE, of our proposed model, DST, against several baseline models across four datasets: Air pollution, Carbon, Electricity, and Weather. The observed variability in MSE across runs primarily stems from stochastic elements in the training process, including random weight initialization, and data shuffling due to mini-batch sampling.
One baseline model, TimesFM, is excluded from the plots because it is evaluated solely in the zero-shot inference setting. Consequently, it produces the MSE value that is reported in Table~\ref{tab2}, with no variability.
This evaluation is conducted on a multivariate input-multivariate output forecasting task, where the input and output lengths are set to 336 and 96, respectively. The results demonstrate that DST consistently achieves lower MSE values compared to the baselines, indicating its superior forecasting capability. Transformer-based models, such as Autoformer and PatchTST, exhibit relatively higher median errors and greater variability. DLinear and s-Mamba perform relatively well, but are still outperformed by DST in all datasets. We note that the y-axis is exaggerated in some cases to better highlight performance differences between models.

\section{Ablation Study}
\label{sec:ablation}
We conducted an ablation study to evaluate the effectiveness of each module in DST to better explain its performance.
The modules under evaluation are (1) Decomposition, (2) Date and Time Embedding, (3) Temporal Feature Extraction, and (4) Spatial Feature Extraction. For each dataset, we compare the model's performance without each of these modules.
Due to the modular structure of DST, we can easily remove individual modules without impacting the functionality of the remaining components. The only exception is the Date and Time Embedding module, whose output is concatenated with the extracted spatio-temporal features. To remove the influence of this module, we replace its output with a zero vector.

Table~\ref{tab5} demonstrates the effectiveness of each module within the model architecture. Among all the components, the temporal feature extraction module plays the most influential role, as indicated by the highest increase in MSE when this module is removed. This suggests that capturing temporal dependencies is crucial for accurately modeling the datasets, particularly for data characterized by strong temporal patterns, such as air pollution, carbon emissions, and electricity usage. Decomposition is also among the most effective components, especially for air pollution and weather forecasting.
The ablation results also reveal that other modules, i.e. date and time embedding and spatial feature extraction, contribute significantly to model performance, albeit to a lesser extent. These components work synergistically to enhance predictive accuracy by providing complementary information that captures different aspects of the data, such as spatial correlations and temporal variations. Overall, the study reveals the necessity of taking a multi-faceted approach that integrates various techniques to effectively handle complex, real-world datasets.

Additionally, we carried out experiments to evaluate the impact of the number of GATv2 layers on DST's performance. We found that across all experiments, a single layer for spatial feature extraction consistently yields the best performance. As a result, DST is implemented with a single GATv2 layer.

\begin{table}[!t]
\small
\setlength{\tabcolsep}{3pt} 
\centering
\begin{tabular}{l c c c c}
\toprule
\textbf{Dataset} & 
\textbf{Air Pollution} &
\textbf{Carbon} &
\textbf{Electricity} &
\textbf{Weather}\\ 
\midrule
DST w/o Decomposition &0.441 &0.474 &0.154 & 0.179\\ 
DST w/o Date\&Time Emb. &0.424 &0.482 &0.149 & 0.158\\
DST w/o Temporal F.E. &0.452 &0.491 &0.162 & 0.161\\
DST w/o Spatial F.E. &0.439 &0.488 &0.156 & 0.165\\
\midrule
DST &\textbf{0.404} &\textbf{0.460} &\textbf{0.133} &\textbf{0.155}\\

\bottomrule
\end{tabular}
\caption{Ablation studies on the effectiveness of key modules. Reported values are the MSE of each setting.}\label{tab5}
\end{table}

\section{Conclusion}
In this paper, we introduced a novel multivariate time-series forecasting model that leverages a GNN architecture integrating GATv2 and TCN. 
This GNN relies on a distinct graph structure per time-series component. 
To learn this graph structure, we implemented a preprocessing step to decompose the time-series into its fundamental components and proposed an effective algorithm for learning the per-component graph structure, based on historical sensor data.
Extensive experiments on four real-world urban datasets with different temporal resolutions, numbers of sensors, and interdependencies confirm that DST consistently achieves the best long-term forecasting performance across different forecast horizons. 
Specifically, it outperforms 
the best-in-class time-series forecasting model by 2.89\% to 9.10\% in different datasets.

Future work will focus on using these forecasting models in various decision-making tasks in the urban environment, from infrastructure planning to assignment of AI workloads (particularly training and fine-tuning) to datacenters, to quantify improvements in decision quality resulting from more accurate long-term forecasts.
Furthermore, we intend to explore how training the long-term forecasting models using the task loss would change the decision quality.

\bibliographystyle{unsrtnat}


\end{document}